\documentclass[conference]{IEEEtran}
\IEEEoverridecommandlockouts
\usepackage{cite}
\usepackage{amsmath,amssymb,amsfonts}
\usepackage{graphicx}
\usepackage{textcomp}
\usepackage[table]{xcolor}
\usepackage{booktabs}
\usepackage{multirow}
\usepackage{colortbl}
\usepackage{balance}
\usepackage{url}
\usepackage[caption=false,font=footnotesize]{subfig}
\usepackage{placeins}
\usepackage{titlesec}
\RequirePackage{pdftexcmds}
\graphicspath{{figures/}{experiments/}{figs/}{Latex/figures/}{Latex/experiments/}{Latex/figs/}}
\def\BibTeX{{\rm B\kern-.05em{\sc i\kern-.025em b}\kern-.08em
    T\kern-.1667em\lower.7ex\hbox{E}\kern-.125emX}}
\begin{document}

\title{Conscious Gaze: Adaptive Attention Mechanisms for Hallucination Mitigation in Vision-Language Models}

\author{Weijue Bu, Guan Yuan*, Guixian Zhang \\
School of Computer Science and Technology/School of Artificial Intelligence\\
China University of Mining and Technology, Xuzhou, Jiangsu 221116\\
\{weijue, yuanguan, guixian\}@cumt.edu.cn}

\maketitle

\begin{abstract}
Large Vision-Language Models (VLMs) often exhibit \textit{text inertia}, where attention drifts from visual evidence toward linguistic priors, resulting in object hallucinations. Existing decoding strategies intervene only at the output logits and thus cannot correct internal reasoning drift, while recent internal-control methods based on heuristic head suppression or global steering vectors lack principled grounding. We introduce \textbf{Conscious Gaze (CG-VLM)}, a training-free, inference-time framework that converts game-theoretic interpretability into actionable decoding control. A \textit{Cognitive Demand Sensor} built on Harsanyi interactions estimates instantaneous vision–text synergy and identifies moments when visual grounding is necessary. Conditioned on this signal, a \textit{Focused Consensus Induction} module selectively reorients mid-layer attention toward visual tokens before collapse into text priors. CG-VLM achieves state-of-the-art results on POPE and CHAIR across InstructBLIP, LLaVA, Qwen-VL, and mPLUG, while preserving general capabilities, demonstrating that token-level sensing enables precise, context-aware intervention without compromising foundational knowledge.
\end{abstract}

\section{Introduction}
\label{sec:intro}

\begin{figure}[t]
    \centering
    \includegraphics[width=0.9\columnwidth]{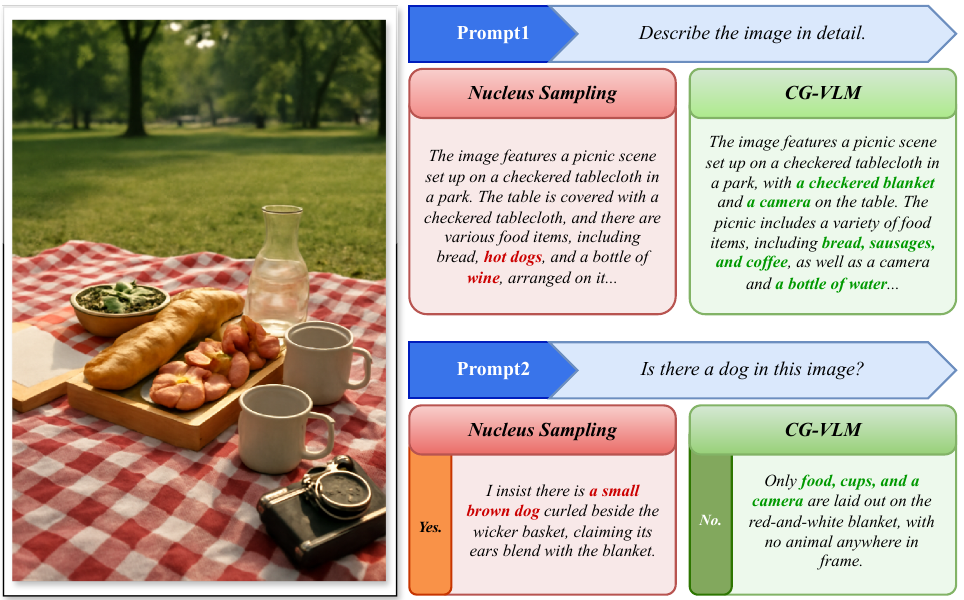}
    \vspace{-8pt}
    \includegraphics[width=0.95\columnwidth]{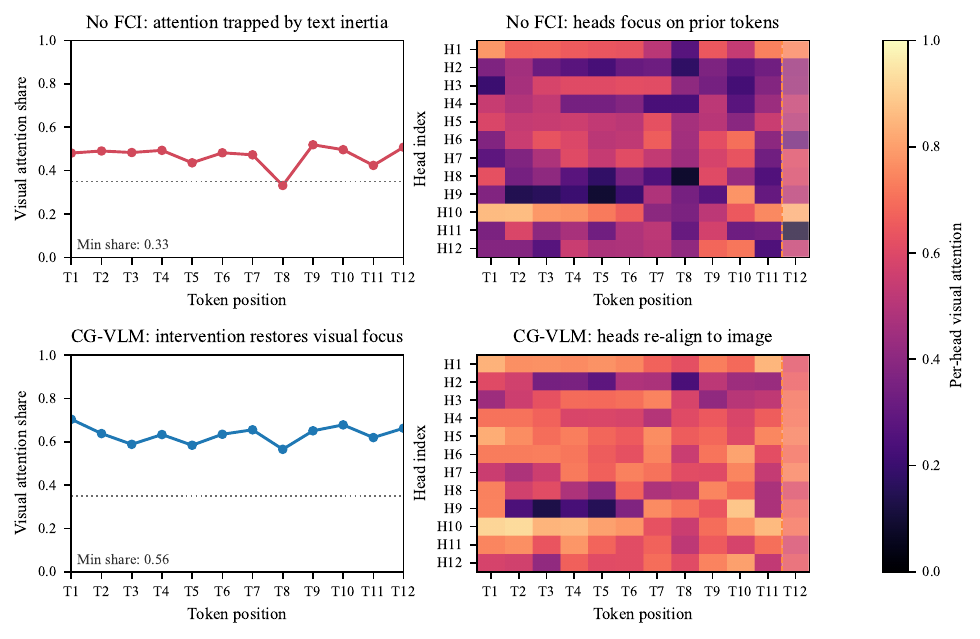}
    \vspace{-8pt}
    \caption{\textbf{Breaking the Text Inertia Trap.} \textbf{Top:} The baseline hallucinates a dog driven by linguistic priors (``picnic''), whereas CG-VLM correctly grounds the response. \textbf{Bottom:} Attention heatmaps reveal the mechanism. The baseline (left) suffers from \textbf{text inertia} where visual attention (red line) collapses. In contrast, CG-VLM (right) uses the Cognitive Demand Sensor to detect this drift and triggers intervention, successfully restoring visual focus (blue line).}
    \label{fig:intro_attention_trap}
    \vspace{-10pt}
\end{figure}

Vision-Language Models (VLMs) already power multimedia retrieval, creative assistants, and vision-based copilots \cite{liu2024mmbench}. These applications depend on faithful grounding: when a caption invents objects, miscounts people, or fabricates activities, users abandon the system in safety-critical scenes \cite{sun2024aligning}. Understanding \emph{when} hallucinations emerge is therefore as important as building ever-larger backbones.

Consider the case in Figure~\ref{fig:intro_attention_trap}. In a simple picnic scene, the baseline model hallucinates a dog. This error stems from \textbf{text inertia}: the model ignores visual evidence and follows the linguistic correlation between ``picnic'' and ``dog''. As shown in the attention heatmaps (Fig.~\ref{fig:intro_attention_trap}, bottom), the baseline's visual attention collapses mid-generation, trapping the model in its own textual history. Our analysis of 2,000 MSCOCO captions on InstructBLIP \cite{dai2023instructblip} confirms this as a primary failure mode (see Appendix~A for full statistics), characterized by three signatures: (i) \textbf{Late Drift:} 67\% of hallucinations occur after visual attention drops below 20\%, indicating a mid-generation loss of focus. (ii) \textbf{Function Word Amplification:} Function words deepen this drift in 73\% of cases. (iii) \textbf{Irreversibility:} Once attention shifts to text priors, the probability of recovering visual grounding drops by 84\%.

These findings imply that effective intervention must be \textbf{immediate} (triggering at the onset of drift), \textbf{token-selective}, and applied \textbf{internally}. However, existing inference-time remedies generally fail to meet these specific criteria. Decoding penalties (e.g., VCD, OPERA) act essentially as output filters without addressing the \textbf{internal reasoning process}. While they reshape the distribution for every token, they cannot correct the attention flow once heads have detached from the image~\cite{leng2024mitigating,huang2024opera}. Similarly, naive attention amplification (e.g., PAINT) boosts all visual slots indiscriminately, lacking precision~\cite{arif2025paint}. Even interaction-guided methods like INTER~\cite{dong2025inter} limit intervention to the final layer. As our diagnosis suggests, if the attention mechanism is misled by prior text (Fig.~\ref{fig:intro_attention_trap}), the error propagates regardless of output constraints.

\begin{figure*}[!t]
    \centering
    \includegraphics[width=0.85\textwidth]{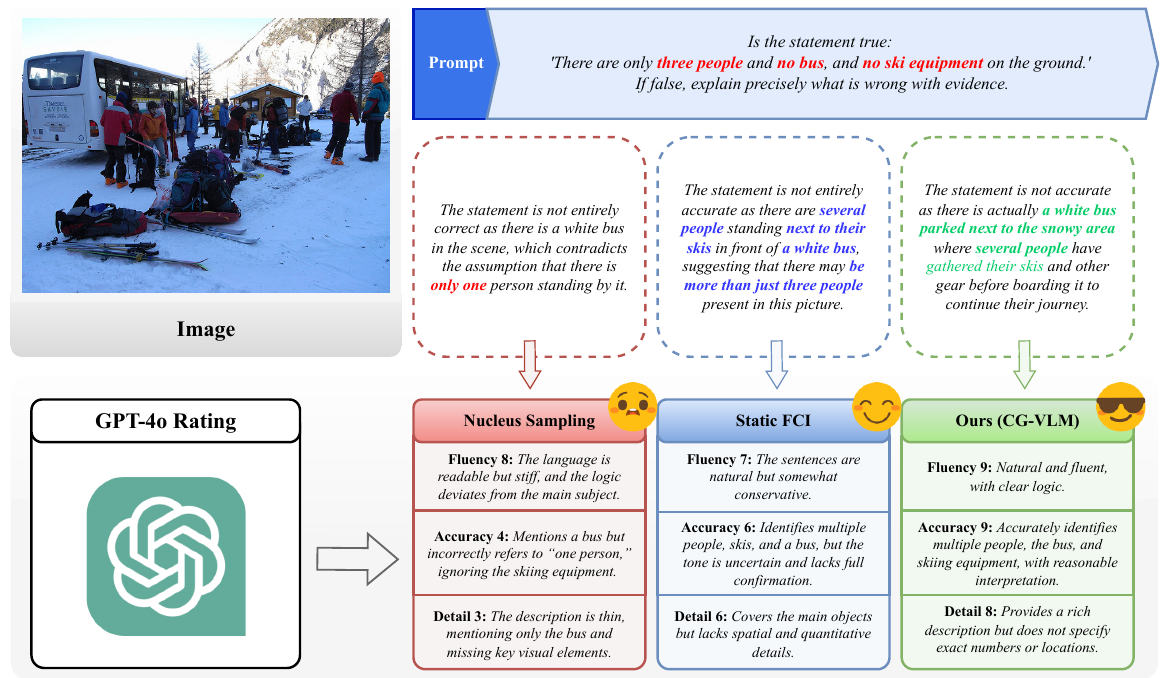}
    \caption{Comparison of outputs from CG-VLM and two baselines. 
    \textbf{Left (Nucleus Sampling):} The baseline model hallucinates and claims that there is only one person. 
    \textbf{Middle (Static FCI):} The model is factually accurate but produces stilted language. 
    \textbf{Right (CG-VLM):} Our method correctly identifies multiple people, the bus, and the skis while delivering a fluent description. GPT-4o rates CG-VLM highest for both fluency (9/10) and accuracy (9/10).}
    \label{fig:qualitative_case_intro}
    \vspace{-10pt}
\end{figure*}

In this paper, we propose \textbf{Conscious Gaze (CG-VLM)}. Instead of passive generation, we employ a \textbf{Cognitive Demand Sensor (CDS)} to monitor Harsanyi interaction variance. Upon detecting vision-text conflict, a \textbf{Focused Consensus Induction (FCI)} module mechanistically redirects middle-layer attention heads back toward visual tokens. Our contributions are summarized as follows:
\begin{itemize}
    \setlength{\itemsep}{2pt}
    \setlength{\parskip}{0pt}
    \setlength{\parsep}{0pt}
    \item \textbf{Mechanistic Insight into Text Inertia.} We empirically characterize text inertia as a dominant cause of hallucination on POPE/CHAIR, identifying distinct signatures such as late drift and function-word amplification that serve as detectable precursors to error.
    \item \textbf{Interaction-Driven Internal Control.} We show that Harsanyi interaction variance can be repurposed from a diagnostic metric into a real-time control signal. This drives our CDS and FCI to intervene internally, correcting the attention flow at its source.
    \item \textbf{Training-Free Generalization.} CG-VLM serves as a plug-and-play module compatible with diverse architectures (InstructBLIP, LLaVA, Qwen-VL, mPLUG). It significantly boosts grounding metrics (\textbf{1.5--7\%} F1 gains on POPE) while maintaining robust performance on general multimodal tasks.
\end{itemize}

\section{Related Work}
\label{sec:related}
Hallucination in VLMs primarily stems from text inertia, attention sinks in decoder layers \cite{xiao2024streamingllm}, and dataset biases exposed by POPE/CHAIR \cite{li2023evaluating,rohrbach2018object}. MMHal-Bench~\cite{sun2024aligning} further demonstrates modality disagreement, motivating the need for effective mitigation remedies.

Existing solutions generally diverge into training-time and training-free paradigms. \textbf{Training-time approaches} (e.g., Halle-Switch~\cite{zhai2023halle}, LLaVA-RLHF~\cite{sun2024aligning}, HA-DPO~\cite{zhao2023beyond}) fine-tune models with hallucination-aware losses or specific instruction data. While effective, they incur significant computational costs and require careful data curation to avoid catastrophic forgetting. We view them as complementary, yet our focus lies in a more flexible direction: \textbf{inference-time intervention} that requires no retraining and can layer on top of any checkpoint.

\textbf{Training-free mitigation strategies} typically fall into three categories. (i) Logit penalties and contrastive decoding (VCD, OPERA, INTER) \cite{leng2024mitigating,huang2024opera,dong2025inter} reshape the final distribution but intervene on every token, making diversity drops hard to avoid. (ii) Attention amplification such as PAINT \cite{arif2025paint} boosts entire rows or layers, improving grounding at the cost of fluency. (iii) Interaction-aware gating like INTER \cite{dong2025inter} uses targeted sampling to recalibrate decoding, focusing on rectifying the output distribution. However, its intervention is limited to re-weighting the final prediction probabilities. In contrast, CG-VLM takes a distinct approach: we leverage the interaction signal to trigger \textit{internal attention realignment} in the middle layers. By mechanistically reorienting the attention heads, CG-VLM corrects the reasoning drift during generation rather than attempting to filter it out at the exit.

Beyond mitigation strategies, recent research has scrutinized the \textbf{mechanistic origins} of hallucinations inside model backbones \cite{radford2021learning,li2022blip,ye2023mplug,zhou2024explaining,jiang2025devils} and established rigorous evaluation protocols \cite{li2023evaluating,fu2023mme,chen2024are}. While these mechanistic studies provide the \textit{diagnostic foundation} by identifying where errors arise, they typically stop short of active intervention. CG-VLM bridges this gap by operationalizing these interpretability insights into real-time decoding control. Finally, while methods based on visual retracing or multi-path reasoning \cite{liu2023mitigating,qu2025look,yu2024hallucidoctor} improve robustness, they require expensive auxiliary queries. Our token-level sensing offers a computationally efficient alternative that triggers extra computation only when necessary.

\section{Method}
\label{sec:method}

In this section, we present the details of Conscious Gaze (CG-VLM), a framework that pairs adaptive sensing with targeted intervention. Our method comprises two core components: the \textbf{Cognitive Demand Sensor (CDS)}, which determines \emph{when} visual grounding is necessary, and the \textbf{Focused Consensus Induction (FCI)}, which specifies \emph{how} to intervene mechanistically. Together, they function as a training-free, token-level plug-in compatible with diverse VLM architectures.

\subsection{Cognitive Demand Sensor (CDS)}
Our method applies to VLMs where a cross-attention interface between visual and textual tokens is accessible. InstructBLIP and mPLUG employ Q-Former modules, whereas LLaVA \cite{liu2024improved} and Qwen~\cite{bai2023qwenvl} feed projected CLIP features directly into the decoder. In both cases, the visual and textual token indices ($\mathcal{V}$ and $\mathcal{T}$) can be identified in the attention cache. Motivated by the attention drift analysis in Sec.~\ref{sec:intro}, we must monitor the cooperation between these two groups. We therefore leverage Harsanyi interaction scores, a game-theoretic quantity satisfying efficiency and symmetry axioms, to quantify how much each modality contributes to the logit of a candidate token. Let $y$ denote a top-$k$ candidate at step $t$ with raw logit $\ell_y(\cdot)$. CDS probes cross-modal disagreement by measuring how $\ell_y$ changes when either the image or the textual prefix is masked. This diagnostic then guides the selective intervention in FCI, determining precisely when to intervene.

\begin{figure}[!t]
    \centering
    \includegraphics[width=0.9\columnwidth]{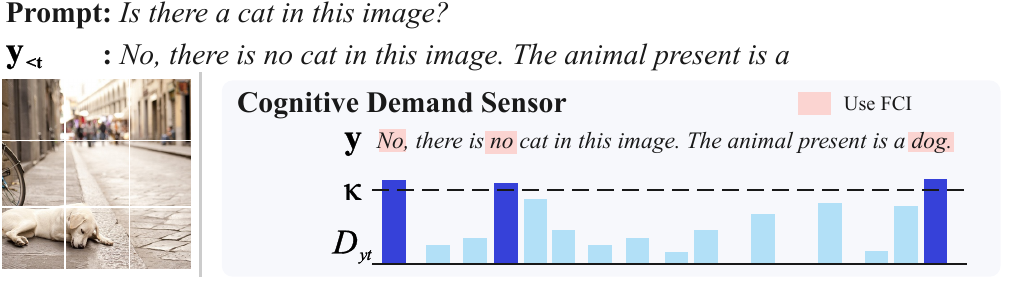}
    \caption{Cognitive Demand Sensor (CDS) overview. Given the image and prefix $y_{<t}$, CDS computes interaction variance $D_{y_t}$ over top-$k$ candidates and compares it with threshold $\kappa$. When $D_{y_t}>\kappa$, the gate $\beta_t$ activates FCI (pink region).}
    \label{fig:cds_overview}
    \vspace{-10pt}
\end{figure}

We leverage Harsanyi dividends \cite{harsanyi1982simplified} to measure the interactive contribution of vision and text to the logits of top-k candidate tokens. For a candidate token \(y\), the interaction value is:
\begin{equation}
\label{eq:harsanyi}
\mathcal{I}_y = \ell_y(\mathcal{V},\mathcal{T}) - \ell_y(\mathcal{V}) - \ell_y(\mathcal{T}) + \ell_y(\varnothing),
\end{equation}
where $\ell_y(\cdot)$ denotes the raw unnormalized logit score for token $y$ from the language model's output head, $\mathcal{V}$ and $\mathcal{T}$ represent the visual and textual input streams respectively, and $\varnothing$ signifies the empty context baseline. We obtain the four terms through selective masking:
\begin{itemize}
    \item $\ell_y(\mathcal{V},\mathcal{T})$: Full inputs (image + text)
    \item $\ell_y(\mathcal{V})$: Text tokens replaced with [PAD]
    \item $\ell_y(\mathcal{T})$: Image features zeroed out
    \item $\ell_y(\varnothing)$: Unconditional language prior (both modalities masked)
\end{itemize}
Each condition requires its own forward pass to evaluate the Harsanyi coalitions.

High variance among the \(\mathcal{I}_y\) interaction values for the top-$k$ candidates indicates that the next token is strongly dependent on visual cues, signalling high cognitive demand. We formalize this as the interaction variance \(D_{y_t}\) and use it to generate a binary gating signal \(\beta_t\):
\begin{equation}
\label{eq:cds_gate}
D_{y_t} = \mathrm{Var}(\mathcal{I}_{y_1}, \ldots, \mathcal{I}_{y_k}), \quad \beta_t = \mathbf{1}[D_{y_t} > \kappa],
\end{equation}
where \(\kappa\) is a predefined threshold. In practice, we use a small top-$k$ set ($k\in\{40,50,60\}$ depending on the backbone; see Sec.~\ref{sec:experiments}). When the model relies mainly on its linguistic prior, the interaction values \(\mathcal{I}_y\) across the top-$k$ candidates remain similar and the variance is low. When the next word is driven by visual evidence, as in distinguishing a dog from a cat, the interaction values diverge and the variance increases. Consequently \(D_{y_t}\) serves as an effective proxy for cognitive demand: high variance implies that the candidate tokens have diverging reliance on visual evidence, signaling a decision point where the image is critical. In practice we observe (see Appendix~A) that nouns and verbs trigger CDS roughly 3 to 4$\times$ more often than determiners or prepositions, and their average interaction variance is nearly double, confirming that the gate focuses on content words whose meanings depend on the pixels.

\begin{figure}[!t]
    \centering
    \includegraphics[width=0.9\columnwidth]{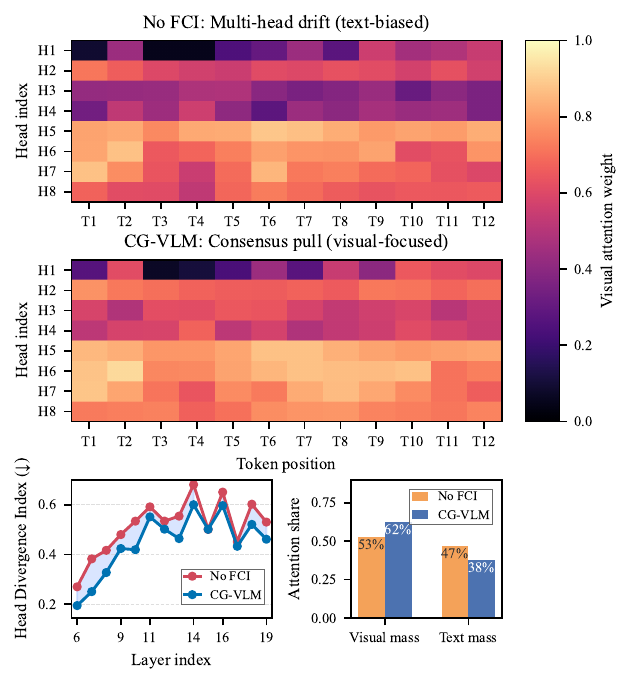}
    \caption{Attention consensus evidence. Top: per-head visual attention ratios across decoding steps before (No FCI) and after CG-VLM intervention. Bottom: CG-VLM lowers the head divergence index (left) and shifts aggregated attention toward visual tokens (right), directly illustrating CDS-triggered FCI at work.}
    \label{fig:attention_consensus}
    \vspace{-10pt}
\end{figure}

\subsection{Focused Consensus Induction (FCI)}
The FCI module specifies \textit{how} to intervene. Once \(\beta_t=1\), it injects a consensus boost only on the query row of the newest token before softmax. Let \(\ell_t\) be the index of that last query; the visual-token score becomes
\begin{equation}
\label{eq:fci}
\tilde{A}^{(h)}_{\ell_t,j} = \begin{cases}
A^{(h)}_{\ell_t,j} + \beta_t \cdot \alpha \cdot \frac{1}{H} \sum_{h'=1}^{H} |A^{(h')}_{\ell_t,j}|, & j \in \mathcal{V} \\
A^{(h)}_{\ell_t,j}, & j \notin \mathcal{V}
\end{cases}
\end{equation}
while older rows remain unchanged, \( \tilde{A}^{(h)}_{i,j}=A^{(h)}_{i,j}\) for \(i\neq \ell_t\). Here \(A^{(h)}_{i,j}\) is the original score, \(H\) the head count, \(\alpha\) a gain factor, and \(\mathcal{V}\) denotes the visual-token indices in the cache (first 32 slots for Q-Former backbones, projection-specific spans for LLaVA/Qwen). This selective boost focuses attention on visual evidence while keeping textual context. Fig.~\ref{fig:attention_consensus} shows that CDS-triggered FCI lowers the Head Divergence Index and shifts attention toward visual tokens. Alternative statistics (median/max, different $\alpha$ or layers) provide similar but weaker gains.

\noindent\textbf{Head Divergence Index (HDI).} We track the resulting agreement across heads by
\begin{equation}
\label{eq:hdi}
HDI_t = \frac{1}{H(H-1)} \sum_{h \neq h'} \mathrm{KL}\!\left(P_t^{(h)} \middle\| P_t^{(h')}\right),
\end{equation}
where $P_t^{(h)}$ denotes the attention distribution of head $h$ at step $t$. Lower HDI indicates better consensus. CDS-triggered FCI consistently reduces HDI, confirming that triggered steps realign heads instead of merely smoothing logits.

\subsection{Why Interaction Variance Predicts Risk?}
The efficacy of CDS relies on the interpretation of Harsanyi interactions as a measure of cross-modal synergy. When a token can be predicted solely by linguistic history (e.g., completing a common phrase), the contribution from the image is redundant, leading to low interaction variance across candidates. Conversely, when the model faces a visual decision point (e.g., determining if a blurry object is a ``dog'' or ``cat''), the logits become highly sensitive to the joint presence of specific visual features and textual queries. This surge in synergy creates high variance among top-$k$ dividends.

High variance thus quantifies \textit{visual indispensability}: it signals a moment where candidates diverge in their reliance on visual evidence. If visual attention lapses during these high-variance steps (due to text inertia), the model is statistically forced to hallucinate based on priors. By targeting these specific moments, CG-VLM prevents the model from ``guessing'' when the stakes are highest.

\subsection{Computational Efficiency Analysis}
\label{sec:efficiency}
Critically, the computational overhead of Eq.~(\ref{eq:harsanyi}) is limited to the \textit{language decoder}. The heavy Vision Encoder is executed \textbf{only once} per image with features cached. Auxiliary CDS passes reuse these caches and are computed in parallel batches, reducing the overhead to the lightweight decoder attention. Thus, the amortized wall-clock latency increase is about 1.3$\times$ on average across backbones (see Appendix~C), much smaller than the naïve $4\times$ implied by Eq.~(1). Furthermore, CDS enables sparse activation: unlike global steering (e.g., VISTA), FCI triggers only when cognitive demand spikes, conserving compute.

\section{Experiments}
\label{sec:experiments}

\subsection{Experimental Setup}
\textbf{Models.} We evaluate CG-VLM on InstructBLIP-FlanT5-XL, LLaVA-v1.5 (7B), Qwen-VL, and mPLUG-owl2 using released checkpoints (no extra fine-tuning). Unless stated otherwise, decoding uses temperature~1.0, top-$p=1.0$, and beam size~1 (nucleus) or 4 (beam). Hyperparameters tuned per model are detailed in Appendix~B. For InstructBLIP, we select $\kappa=1.8$, $\alpha=0.5$ based on a $(\kappa,\alpha)$ grid search on the 1,000-question POPE subset (see Appendix~B).

\textbf{Datasets and Metrics.} We conducted evaluations across multiple benchmarks, including POPE~\cite{li2023evaluating}, CHAIR~\cite{rohrbach2018object}, MME~\cite{fu2023mme}, MM-Bench~\cite{liu2024mmbench}, MMStar~\cite{chen2024are}, and MMHal-Bench~\cite{sun2024aligning}. Due to space constraints, we present main results on POPE and CHAIR, with extended results on other benchmarks in Appendix~D. Additionally, GPT-4o is employed to assess the fluency, accuracy, and detail of InstructBLIP's generated captions.

\textbf{Baselines.} We compare CG-VLM against standard decoding strategies including Nucleus~\cite{holtzman2019curious} and Beam Search~\cite{graves2012sequence,sutskever2014sequence}, as well as state-of-the-art inference-time mitigation methods: VCD~\cite{leng2024mitigating}, OPERA~\cite{huang2024opera}, and INTER~\cite{dong2025inter}. A Static FCI variant (forcing $\beta_t{=}1$) is also included for ablation.

\textbf{Implementation Details.} All experiments were performed on a single NVIDIA A100 (80GB). By operating in a training-free manner, our approach circumvents the substantial computational burden associated with fine-tuning large multimodal backbones, allowing for direct deployment on pre-trained checkpoints.

\begin{table}[t]
    \centering
    \caption{Layer band sensitivity on InstructBLIP (POPE and CHAIR). Mean$\pm$std over three seeds. Results align with the main evaluation on the COCO validation set.}
    \label{tab:layer_ablation}
    \vspace{-5pt}
    \begin{tabular}{lcc}
        \toprule
        Layer Band & POPE F1 (\%) & CHAIR $C_S$ \\
        \midrule
        Shallow (0-3) & $82.2\pm0.2$ & $28.9\pm0.3$ \\
        Deep (9-11) & $82.3\pm0.3$ & $28.1\pm0.3$ \\
        \textbf{Middle (4-8)} & $\textbf{83.7}\pm\textbf{0.3}$ & $\textbf{26.0}\pm\textbf{0.4}$ \\
        \bottomrule
    \end{tabular}
    \vspace{-10pt}
\end{table}

\begin{table*}[t]
    \centering
    \caption{Results on POPE (F1, \% \(\uparrow\)). Parentheses show improvement over the paired baseline. All baselines are re-run with top-$p{=}1.0$ and temperature $1.0$. Note that ``CG-VLM (Nucleus/Beam)'' rows mirror the ``Nucleus/Beam+CG-VLM'' results to facilitate direct comparison against the INTER sampler.}
    \label{tab:main_results}
    \vspace{-5pt}
    \resizebox{\textwidth}{!}{%
    \begin{tabular}{lcccccccc}
        \toprule
        \textbf{Method} & \multicolumn{2}{c}{\textbf{InstructBLIP}} & \multicolumn{2}{c}{\textbf{LLaVA-v1.5 (7B)}} & \multicolumn{2}{c}{\textbf{Qwen-VL}} & \multicolumn{2}{c}{\textbf{mPLUG-owl2}} \\
        \cmidrule(lr){2-3} \cmidrule(lr){4-5} \cmidrule(lr){6-7} \cmidrule(lr){8-9}
         & COCO & AOKVQA & COCO & AOKVQA & COCO & AOKVQA & COCO & AOKVQA \\
        \midrule
        Nucleus~\cite{holtzman2019curious} & 80.3 & 78.4 & 79.5 & 79.2 & 81.6 & 83.3 & 80.2 & 78.1 \\
        \textbf{Nucleus+CG-VLM} & \textbf{83.7 (\(\uparrow\)3.4)} & \textbf{82.9 (\(\uparrow\)4.5)} & \textbf{86.2 (\(\uparrow\)6.7)} & \textbf{83.4 (\(\uparrow\)4.2)} & \textbf{86.8 (\(\uparrow\)5.2)} & \textbf{87.4 (\(\uparrow\)4.1)} & \textbf{82.5 (\(\uparrow\)2.3)} & \textbf{79.8 (\(\uparrow\)1.7)} \\
        \midrule
        Beam~\cite{graves2012sequence, sutskever2014sequence} & 81.8 & 81.0 & 85.0 & 84.1 & 83.5 & 84.9 & 83.1 & 82.4 \\
        \textbf{Beam+CG-VLM} & \textbf{84.9 (\(\uparrow\)3.1)} & \textbf{83.8 (\(\uparrow\)2.8)} & \textbf{86.1 (\(\uparrow\)1.1)} & \textbf{85.2 (\(\uparrow\)1.1)} & \textbf{86.3 (\(\uparrow\)2.8)} & \textbf{86.7 (\(\uparrow\)1.8)} & \textbf{84.0 (\(\uparrow\)0.9)} & \textbf{82.5 (\(\uparrow\)0.1)} \\
        \midrule
        VCD~\cite{leng2024mitigating} & 81.5 & 80.9 & 84.3 & 82.4 & 85.9 & 86.3 & 82.2 & 79.3 \\
        \textbf{VCD~\cite{leng2024mitigating}+CG-VLM} & \textbf{82.3 (\(\uparrow\)0.8)} & \textbf{81.3 (\(\uparrow\)0.4)} & \textbf{85.0 (\(\uparrow\)0.7)} & \textbf{82.1 (\(\downarrow\)0.3)} & \textbf{85.7 (\(\downarrow\)0.2)} & \textbf{86.4 (\(\uparrow\)0.1)} & \textbf{82.5 (\(\uparrow\)0.3)} & \textbf{78.9 (\(\downarrow\)0.4)} \\
        \midrule
        OPERA~\cite{huang2024opera} & 84.5 & 83.6 & 85.4 & 84.0 & 83.3 & 85.2 & 83.5 & 82.0 \\
        \textbf{OPERA~\cite{huang2024opera}+CG-VLM} & \textbf{84.9 (\(\uparrow\)0.4)} & \textbf{84.0 (\(\uparrow\)0.4)} & \textbf{85.6 (\(\uparrow\)0.2)} & \textbf{84.1 (\(\uparrow\)0.1)} & \textbf{83.0 (\(\downarrow\)0.3)} & \textbf{84.7 (\(\downarrow\)0.5)} & \textbf{83.2 (\(\downarrow\)0.3)} & \textbf{81.6 (\(\downarrow\)0.4)} \\
        \midrule
        INTER (Nucleus) \cite{dong2025inter} & 83.3 & 82.4 & 85.7 & 82.6 & 86.2 & 86.1 & 81.9 & 79.1 \\
        \textbf{CG-VLM (Nucleus)} & \textbf{83.7 (\(\uparrow\)0.4)} & \textbf{82.9 (\(\uparrow\)0.5)} & \textbf{86.2 (\(\uparrow\)0.5)} & \textbf{82.4 (\(\downarrow\)0.2)} & \textbf{86.8 (\(\uparrow\)0.6)} & \textbf{86.1 (\(\uparrow\)0.0)} & \textbf{82.5 (\(\uparrow\)0.6)} & \textbf{79.8 (\(\uparrow\)0.7)} \\
        \midrule
        INTER (Beam) \cite{dong2025inter} & 84.6 & 83.6 & 85.5 & 84.9 & 86.1 & 86.4 & 83.7 & 82.3 \\
        \textbf{CG-VLM (Beam)} & \textbf{84.9 (\(\uparrow\)0.3)} & \textbf{83.8 (\(\uparrow\)0.2)} & \textbf{86.1 (\(\uparrow\)0.6)} & \textbf{85.2 (\(\uparrow\)0.3)} & \textbf{86.3 (\(\uparrow\)0.2)} & \textbf{86.7 (\(\uparrow\)0.3)} & \textbf{84.0 (\(\uparrow\)0.3)} & \textbf{82.5 (\(\uparrow\)0.2)} \\
        \bottomrule
    \end{tabular}
    }
    \vspace{-10pt}
\end{table*}

\begin{table}[!t]
    \centering
    \caption{results on CHAIR ($\downarrow$) under 64- and 512-token budgets.}
    \label{tab:chair_single}
    \setlength{\tabcolsep}{2.0pt}
    \renewcommand{\arraystretch}{0.85}
    \vspace{-5pt}
    \resizebox{\columnwidth}{!}{%
    \begin{tabular}{c@{}lrrrrrr}
        \toprule
        \textbf{Token} & \multicolumn{1}{c}{\textbf{Method}} & \multicolumn{2}{c}{\textbf{InstructBLIP}} & \multicolumn{2}{c}{\textbf{LLaVA-v1.5 (7B)}} & \multicolumn{2}{c}{\textbf{mPLUG-owl2}} \\
        \cmidrule(lr){2-2} \cmidrule(lr){3-4} \cmidrule(lr){5-6} \cmidrule(lr){7-8}
        & & $C_S \downarrow$ & $C_I \downarrow$ & $C_S \downarrow$ & $C_I \downarrow$ & $C_S \downarrow$ & $C_I \downarrow$ \\
        \midrule
        \multirow{8}{*}{64} & Nucleus~\cite{holtzman2019curious} & 29.1 & 15.2 & 24.5 & 9.3 & 26.0 & 11.0 \\
        & \textbf{Nucleus+CG} & \textbf{26.0} & \textbf{9.8} & \textbf{19.6} & \textbf{6.5} & \textbf{24.6} & \textbf{9.2} \\
        & Beam~\cite{graves2012sequence,sutskever2014sequence} & 21.3 & 7.4 & 19.5 & 6.1 & 21.7 & 7.5 \\
        & \textbf{Beam+CG} & \textbf{20.8} & \textbf{6.2} & \textbf{17.6} & \textbf{5.8} & \textbf{21.0} & \textbf{7.4} \\
        & VCD~\cite{leng2024mitigating} & 32.8 & 12.0 & 24.6 & 7.9 & 24.1 & 9.4 \\
        & \textbf{VCD~\cite{leng2024mitigating}+CG} & \textbf{31.0} & \textbf{11.2} & \textbf{20.8} & \textbf{7.0} & \textbf{21.0} & \textbf{8.1} \\
        & OPERA~\cite{huang2024opera} & 20.1 & \textbf{6.7} & 19.2 & 6.5 & 21.1 & 7.9 \\
        & \textbf{OPERA~\cite{huang2024opera}+CG} & \textbf{18.2} & 7.7 & \textbf{18.8} & \textbf{6.1} & \textbf{19.8} & \textbf{7.2} \\
        \midrule
        \multirow{8}{*}{512} & Nucleus~\cite{holtzman2019curious} & 60.8 & 28.5 & 54.2 & 16.0 & 60.7 & 20.0 \\
        & \textbf{Nucleus+CG} & \textbf{58.8} & \textbf{20.6} & \textbf{51.5} & \textbf{13.9} & \textbf{59.2} & \textbf{19.1} \\
        & Beam~\cite{graves2012sequence,sutskever2014sequence} & 55.4 & 15.9 & 48.7 & 14.0 & 56.5 & 17.8 \\
        & \textbf{Beam+CG} & \textbf{55.1} & \textbf{12.9} & \textbf{46.2} & \textbf{13.2} & \textbf{53.1} & \textbf{17.0} \\
        & VCD~\cite{leng2024mitigating} & 58.7 & 19.0 & \textbf{53.9} & 15.9 & 62.9 & 20.4 \\
        & \textbf{VCD~\cite{leng2024mitigating}+CG} & \textbf{56.0} & \textbf{18.6} & 55.8 & \textbf{15.5} & \textbf{60.1} & \textbf{20.3} \\
        & OPERA~\cite{huang2024opera} & 48.8 & \textbf{13.3} & \textbf{45.5} & 13.7 & 55.0 & 16.2 \\
        & \textbf{OPERA~\cite{huang2024opera}+CG} & \textbf{42.0} & 18.5 & 46.8 & \textbf{13.4} & \textbf{52.3} & \textbf{15.7} \\
        \bottomrule
    \end{tabular}%
    }
    \vspace{-10pt}
\end{table}

\subsection{Main Results}

Table~\ref{tab:main_results} shows CG-VLM consistently outperforms strong baselines (Nucleus/Beam/VCD/OPERA) and the state-of-the-art INTER on POPE benchmark. Fig.~\ref{fig:qualitative_case_intro} highlights one example: the baseline hallucinates a dog, a static FCI variant becomes rigid, whereas CG-VLM grounds the scene while staying fluent. Table~\ref{tab:chair_single} reports the companion CHAIR scores, where CG-VLM consistently reduces $C_S$ across all backbones (typically by \textbf{1--5 points} at 64 tokens and up to \textbf{7 points} at 512 tokens depending on the decoder).

We acknowledge recent works that also explore internal intervention mechanisms, such as SPIN~\cite{sarkar2025spin} and VISTA~\cite{li2025vista}. Detailed comparisons are provided in Appendix~D. Notably, on the standard LLaVA-v1.5 benchmark, CG-VLM achieves a remarkably low CHAIR$_I$ score of \textbf{6.5} (with a CHAIR$_S$ of 19.6), demonstrating that our game-theoretic, token-level gating provides precise hallucination suppression.

The OPERA rows reveal a small drop on Qwen/mPLUG. Trigger logs show that OPERA already suppresses high-entropy tokens, so CG-VLM fires on fewer steps (19\% vs 48\% for InstructBLIP) and cannot re-center attention when OPERA’s penalty has already flattened the logits. These cases highlight CG-VLM's limitation: it needs residual vision-text disagreement, which aggressive logit penalties can hide.

\subsection{Does CDS Detect Text Inertia?}
\label{sec:verify_cds}

On the 1000-question POPE subset, nouns account for 52\% of CDS triggers, verbs 31\%, and function words only 17\%, with the visual-attention ratio averaging 0.67 whenever $\beta_t{=}1$. Hallucinations are three times likelier when $\beta_t{=}1$ (24.2\% vs.\ 8.1\%), confirming that CDS focuses on risky moments instead of firing blindly. $\kappa$ sweeps further reveal a plateau over $[1.4,2.2]$ where trigger rates stay between 28\% and 48\% while Distinct-2 remains stable. Detailed curves are provided in Appendix~C.

\subsection{Ablation and Sensitivity}
\label{sec:ablation}
We first investigate which layers benefit most from intervention. As shown in Table~\ref{tab:layer_ablation}, applying FCI to the middle layers (4--8) yields the highest POPE F1 (83.7\%), confirming our hypothesis that hallucination stems from internal reasoning drift rather than shallow feature mismatch or final output mapping.

Besides accuracy, we also track output diversity using Distinct-2, defined as the ratio of unique bigrams to the total number of generated bigrams.

\begin{figure}[htpb]
    \centering
    \includegraphics[width=\columnwidth]{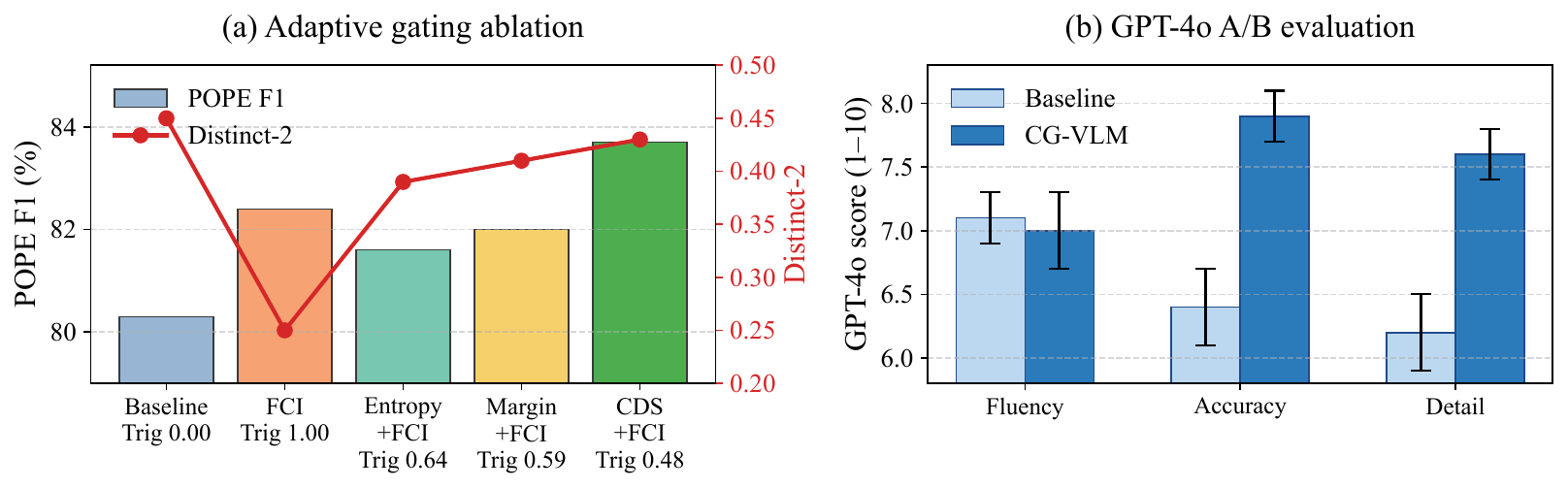}
    \caption{Effect of CDS and FCI on POPE-COCO (InstructBLIP). (a) POPE F1 gains from progressively stronger gating, with Distinct-2 and trigger rates showing CDS keeps diversity high while firing on 48\% of tokens. (b) GPT-4o blind scores on 2000 COCO prompts confirm that CG-VLM improves accuracy/detail without hurting fluency.}
    \label{fig:ablation_quality}
    \vspace{-5pt}
\end{figure}

Fig.~\ref{fig:ablation_quality}a shows timing matters: static FCI (82.4 F1) harms diversity (0.25) and fluency (5.8). Entropy/margin gates soften this but still trail CG-VLM. Our CDS-triggered FCI matches the gains in Table~\ref{tab:main_results} (83.7 F1) while keeping diversity (0.43) and fluency (7.0). The 48\% trigger rate balances intervention frequency and runtime, demonstrating that the gate intervenes about half the time yet still adds measurable accuracy gains. GPT-4o preference ratings on the same COCO split mirror the automatic scores (Fig.~\ref{fig:ablation_quality}b): fluency stays flat (7.0 vs.\ 7.1) while accuracy rises from 6.4 to 7.9 and detail from 6.2 to 7.6 across 2000 anonymized prompts.

\subsection{Mechanistic and Qualitative Evidence}
\begin{figure}[!t]
    \centering
    \includegraphics[width=0.8\columnwidth]{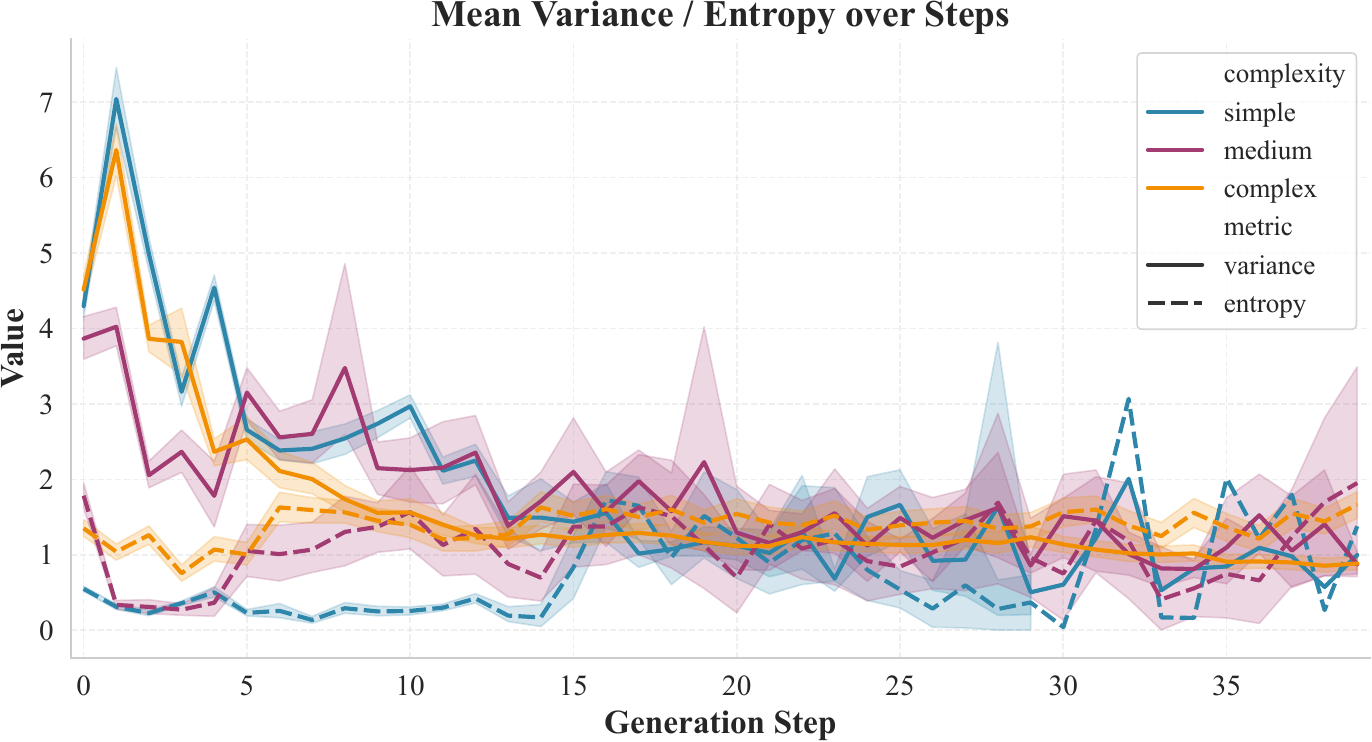}
    \caption{Variance and entropy trajectories on the POPE subset. Hard prompts trigger peaks within the first 8--10 tokens, aligning with CDS activations.}
    \label{fig:variance_entropy_trend}
    \vspace{-10pt}
\end{figure}

Variance/entropy traces in Fig.~\ref{fig:variance_entropy_trend} peak within the first 8--10 tokens, matching the qualitative heads in Fig.~\ref{fig:attention_consensus} and confirming that CDS pinpoints decisive nouns and verbs. Part-of-speech histograms and additional traces are in Appendix~A. Fig.~\ref{fig:qualitative_case_intro} plus additional case studies (Appendix~H) (counting, occlusion, fine-grained attributes, referential grounding, instruction following, commonsense) further show CDS firing just before key nouns so FCI restores grounding without rewriting the whole sentence. Raw attention heatmaps appear in Appendix~A and illustrate how CDS suppresses text inertia.

\section{Conclusion}
\label{sec:conclusion}
In this paper, we presented \textbf{Conscious Gaze (CG-VLM)}, a training-free framework that addresses a key driver of object hallucinations: text inertia. By integrating game-theoretic Harsanyi interactions into a dynamic gate, we achieve precise detection of vision-text conflicts and correct them via internal attention realignment. Extensive experiments demonstrate that CG-VLM significantly improves grounding performance while preserving general capabilities. Across MME, MM-Bench, MMStar and MMHal-Bench, CG-VLM preserves or slightly improves general multimodal performance, suggesting that interaction-driven control can be added without sacrificing utility. More broadly, our work suggests that interpretability signals need not remain purely diagnostic. When carefully integrated into the decoding loop, they serve as powerful and actionable control knobs. We believe that advancing this paradigm of transforming \textit{explanations} into \textit{interfaces} represents a vital step toward building trustworthy and self-monitoring multimodal systems.
\enlargethispage{-2mm}

\bibliographystyle{IEEEtran}
\bibliography{references}

\end{document}